# MULTIMODAL AMBIVALENCE AND HESITANCY RECOGNITION WITH BIDIRECTIONAL AUDIO–TEXT CROSS-ATTENTION

**TingTing Gu**
JINAN University
ttinggu@stu2024.jnu.edu.cn

**JinCheng Luo**
JINAN University
jcluo@stu.jnu.edu.cn

**WenQiu Guo**
Xihua University
wenqiuguo3@gmail.com

**ABSTRACT**

Ambivalence and hesitancy (A/H) are subtle behavioural states that can be expressed through both what a person says and how it is spoken. This paper presents an audio–text framework for video-level A/H recognition. Full-video transcripts and speech waveforms are encoded using pre-trained multilingual BERT and wav2vec 2.0, respectively, and projected into a shared representation space. Two stacked bidirectional cross-attention blocks allow linguistic tokens to attend to acoustic frames and acoustic frames to attend to linguistic context without requiring explicit temporal alignment. The resulting sequences are aggregated through masked mean pooling and combined using concatenation, element-wise products, and absolute differences for binary classification. We evaluate the model on the subject-independent split of the Behavioural Ambivalence/Hesitancy (BAH) dataset. The proposed system achieves 72.19% accuracy, 75.00% precision, 81.13% recall, 77.95% A/H F1-score, and 70.16% Macro-F1 on the test set. These results demonstrate the potential ofjointly modelling linguistic and acoustic information for automatic A/H recognition.code at:https://github.com/DorriG/ECCV

## 1 Introduction

Digital behaviour-change interventions—mobile apps and online coaching programs that support smoking cessation, physical activity, dietary change, or mental-health self-management—increasingly rely on the ability to sense a user's psychological state and adapt their content accordingly. Among the states most predictive of intervention failure is ambivalence/hesitancy (A/H): a user's simultaneous or wavering mix of motivation and reluctance toward change. Left undetected, A/H often leads to disengagement or relapse, so recognizing it automatically and in real time would allow personalized systems to intervene with tailored encouragement rather than a generic script.

Recognizing A/H is nonetheless difficult, particularly when relying on only a subset of the available modalities. First, A/H is not a single, sharply defined emotion; ambivalence and hesitancy tend to manifest through overlapping and hard-to-separate cues, which is why the AH Video Recognition Challenge treats them jointly as one binary label rather than as distinct classes. Second, the evidence for A/H is distributed unevenly across the audio and textual channels and over time: a hesitant pause or a shift in vocal tone may appear only in the audio stream, while a qualifying phrase ("I guess...", "maybe...") may appear only in the transcript, often with little temporal overlap between the two. A model that pools each modality independently and concatenates the result risks averaging away exactly these localized, cross-modal correlations.

While the BAH dataset also provides synchronized video and cropped, aligned faces, in this work we deliberately restrict our model to the audio and textual tracks. Spoken language and its prosodic delivery already carry rich and complementary evidence of hesitancy—hedging words, filler sounds, pauses, and changes in tone—and can be modeled effectively with well-established pretrained encoders (BERT for text, Wav2Vec2 for audio); extending the approach to incorporate the visual/facial channel is left for future work.

The AH Video Recognition Challenge formalizes this problem using the BAH dataset, which contains 1,427 videos (10.60 hours total) collected from 300 participants across Canada responding to a fixed set of questions designed to elicit A/H. The dataset provides raw video, cropped and aligned faces, frame- and video-level A/H annotations, speech-to-text transcripts with timestamps, annotator cues, and participant metadata, split participant-wise into training, validation, and test partitions. The task is to predict, for each video, whether A/H is present (1) or absent (0); performance is measured by the macro-averaged F1 score across both classes on a private test set, with average precision of the positive class reported as a secondary metric. A zero-shot multimodal LLM baseline (Video-LLaVA, vision-only) achieves $P = 0.2827$ on the public test set, underscoring both the difficulty of the task and the room for models that make better use of the audio and text modalities.

In this paper we focus on exploiting the complementary, fine-grained relationship between what a participant says and how they say it. Instead of treating text and audio as two independent sources of evidence that are pooled and concatenated only at the end, we couple a pretrained multilingual BERT text encoder with a pretrained Wav2Vec2 audio encoder through a bidirectional cross-modal attention module: text tokens attend to the audio frames co-occurring with them, and audio frames attend to the text tokens that describe them. This lets the model align, for example, a hedging word with the vocal hesitation that accompanies it, before any pooling discards that correspondence. After cross-attention, we mask-pool each modality into a global vector and construct an explicit relational representation—the two pooled vectors together with their element-wise product and absolute difference—so the classifier receives both modality-specific and modality-shared/complementary signals rather than a plain concatenation.

Our contributions are threefold:

- We propose a bidirectional cross-modal attention architecture for A/H recognition that models fine-grained, token-to-frame correspondences between text and audio in both directions.
- We introduce a four-way relational fusion scheme (text, audio, product, absolute difference) that explicitly separates modality-shared from modality-specific evidence before classification.
- We evaluate the resulting method on the BAH dataset under the AH Video Recognition Challenge protocol.

Because the cross-attention weights indicate which audio frames a given word (or vice versa) relies on, the method also offers a natural path toward the interpretable predictions encouraged by the challenge, such as highlighting when in a video A/H cues emerge and which modality drives the decision.

## 2 Related Work

### 2.1 From affective computing to in-the-wild multimodal analysis

Affective computing was established as the study of systems that can recognize, model, and respond to human affect [1]. Early computational formulations were strongly influenced by two complementary views of emotion. The categorical view associates affect with a small set of prototypical emotions and their observable expressions, as exemplified by Ekman and Friesen's cross-cultural study of facial expressions [2]. The dimensional view instead locates affect in a continuous space; Russell's circumplex model organizes affect primarily along valence and arousal [3]. These views continue to shape contemporary benchmarks, which variously predict discrete expressions, continuous valence–arousal values, or both.

The field has progressively moved from posed, single-modality recordings toward spontaneous and multimodal behaviour. IEMOCAP provided synchronized speech, video, motion capture, and text from dyadic interactions, with both categorical and dimensional annotations [4]. CMU-MOSEI scaled multimodal language analysis to thousands of in-the-wild monologue videos annotated for sentiment and emotion [5]. On the visual side, Aff-Wild2 expanded in-the-wild affect analysis across valence–arousal estimation, expression recognition, and facial action units [6]. Collectively, these datasets shifted the focus from recognizing isolated facial configurations or acoustic patterns to modelling affect as temporally evolving behaviour expressed through interacting linguistic, vocal, and visual channels. They also exposed persistent difficulties caused by spontaneous expression, temporal misalignment, speaker variability, and imbalanced labels.

Ambivalence and hesitancy do not fit cleanly into a basic-emotion taxonomy. They are better understood as higher-order behavioural states related to conflict, uncertainty, and readiness to act. Their recognition therefore extends affective computing from identifying **which emotion** is expressed toward determining whether a person is conflicted or hesitant about a topic or action.

### 2.2 Ambivalence, hesitancy, and behavioural change

Psychological research distinguishes the coexistence of positive and negative evaluations from the subjective experience of conflict. Priester and Petty showed that the positive and negative bases of an attitude jointly determine experienced ambivalence, while also demonstrating that objective opposition and felt ambivalence are not identical constructs [7]. Ambivalence is particularly important in behavioural change. Motivational interviewing was developed around eliciting and resolving clients' mixed motivations; Miller and Rose further distinguished decisional-balance procedures from the evocation of change talk when responding to client ambivalence [8]. This connection motivates automatic A/H recognition for adaptive digital interventions: a system that detects conflict or hesitation could, in principle, adjust how and when it responds rather than treating every utterance as an unambiguous commitment.

Hesitancy also has observable correlates in spontaneous speech. Work on conversational speech annotation found that filled pauses and vowel duration increase with annotated hesitation degree [9]. Controlled perception experiments further showed that filled-pause position, speaking rate, and fundamental frequency jointly influence perceived speaker confidence [10]. These findings support the use of an acoustic stream, but they do not imply that pauses alone define A/H: disfluencies can serve ordinary planning or turn-holding functions, and ambivalence may instead be conveyed lexically, visually, or through disagreement between modalities.

The BAH dataset is the first benchmark designed specifically for subject-based, multimodal A/H recognition in videos [11]. It operationalizes A/H as a binary presence/absence label because ambivalence and hesitancy manifest similarly in practice. Three behavioural experts provide temporal segments, frame- and video-level labels, certainty, and cue categories spanning language, audio, facial expression, body movement, and cross-modal inconsistency. BAH therefore connects the psychological and speech-based literature above with in-the-wild affective computing. Its baseline results also show that A/H remains difficult to recognize and that transcript and audio information can be competitive with visual information [11]. Our work addresses BAH's video-level task using the two modalities most directly related to **what is said** and **how it is said**: the transcript and speech waveform.

### 2.3 Pre-trained speech and language representations

The emergence of large pre-trained Transformers changed affect modelling by reducing reliance on features designed for a single dataset. BERT learns bidirectional contextual language representations from unlabelled text [12], while wav2vec 2.0 learns contextual speech representations from raw audio through masked latent prediction and contrastive learning [13]. These models are relevant to A/H because lexical conflict and acoustic uncertainty may occur at different temporal positions and need not be captured by hand-crafted keyword or prosodic lists. BAH transcripts were generated with multilingual Whisper [14]. Whisper is not optimized in our pipeline: the released full-video transcript is encoded by cased multilingual BERT, while the waveform is independently encoded by wav2vec 2.0. This separation preserves both semantic and non-lexical vocal information.

### 2.4 Multimodal fusion and cross-modal attention

Feature concatenation and decision-level averaging are common fusion baselines, but they provide limited mechanisms for modelling local interactions between heterogeneous sequences. The Multimodal Transformer (MulT) introduced directional cross-modal attention for unaligned language, acoustic, and visual sequences, enabling one modality to retrieve relevant elements from another without explicit temporal alignment [15]. Cross-attention has subsequently been adapted to affective computing. The Joint Multimodal Transformer, for example, uses key-based cross-attention to model inter- and intra-modal relationships for in-the-wild emotion and pain recognition [16].

## 3 The BAH Dataset

The Behavioural Ambivalence/Hesitancy (BAH) dataset is a multimodal video dataset developed for the automatic recognition of ambivalence and hesitancy in digital behavioural-change settings [11]. It contains 1,427 videos collected from 300 participants across Canada, with a total duration of approximately 10.60 hours. Participants recorded themselves through a webcam and microphone while answering seven questions designed to elicit different behavioural responses, including neutral, positive, negative, ambivalent, willing, resistant, and hesitant responses. Since participants were allowed to skip questions, the number of recordings varies across individuals.

BAH provides several complementary sources of information for each recording. These include the original video and audio, cropped and aligned facial images, the full speech transcript, timestamped transcript segments, and participant metadata. The transcripts and spoken-language information were generated using the multilingual Whisper model. The dataset also provides detailed A/H annotations, including temporal A/H segments, frame-level labels, a global video-level label, annotator certainty, and the behavioural cues used during annotation. These cues cover facial expression, body movement, language, audio, and inconsistency between modalities. Three behavioural experts participated in the annotation process. At video level, the task is formulated as binary classification: label 0 indicates that no A/H is present, whereas label 1 indicates that A/H occurs in the video.

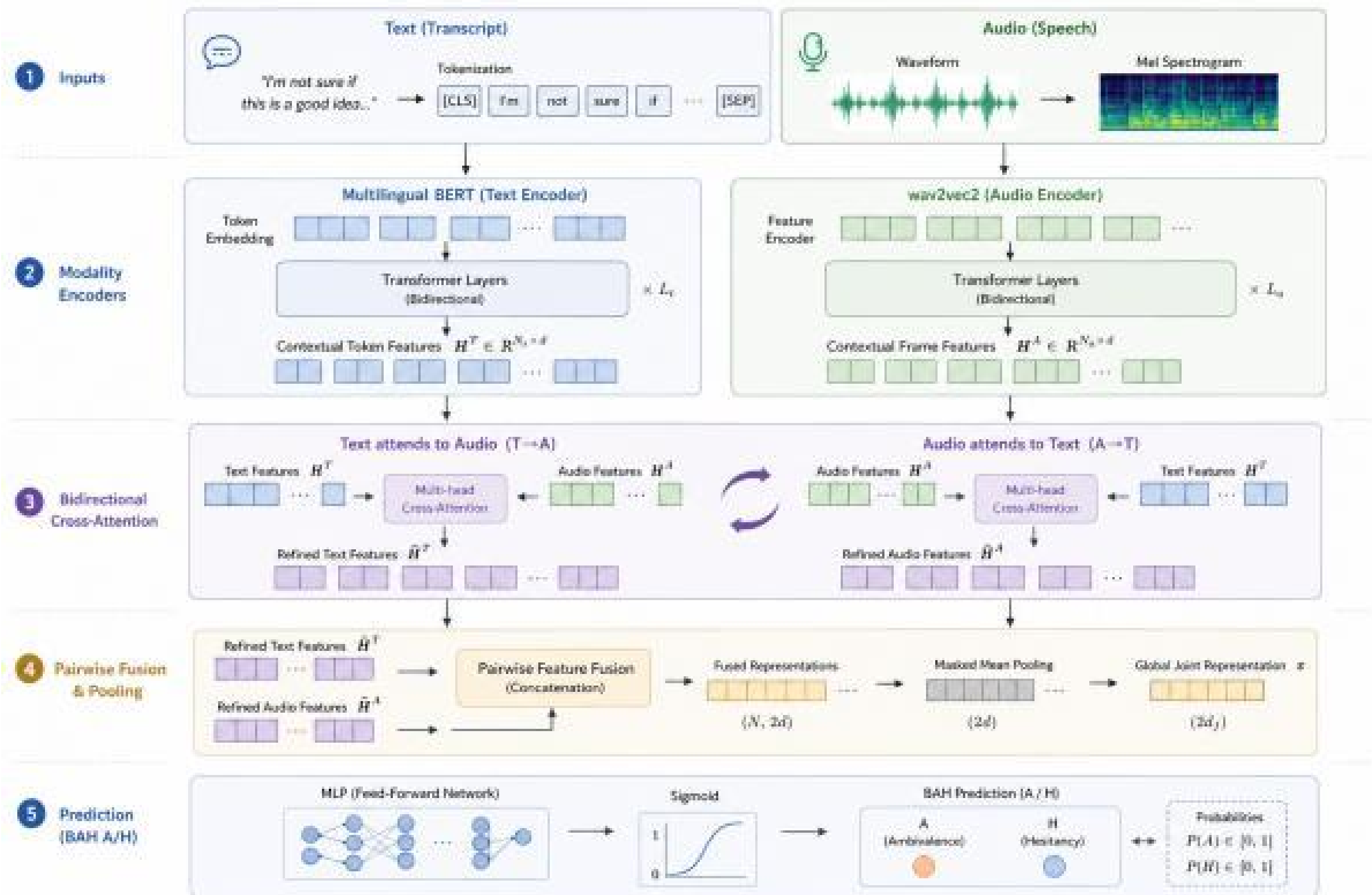

Figure 1: Overview of the proposed audio–text framework for video-level A/H recognition. The transcript and waveform are encoded by multilingual BERT and wav2vec 2.0, projected into a shared space, and exchanged through stacked bidirectional cross-attention blocks. Masked mean pooling and interaction-based fusion produce the final binary prediction.

The official protocol divides the 300 participants into subject-independent training, validation, and test sets, so recordings from the same participant never appear in different subsets. The training set contains 778 videos from 195 participants, the validation set contains 124 videos from 30 participants, and the test set contains 525 videos from 75 participants. Together, the three subsets contain all 1,427 videos in the dataset.

## 4 Method

### 4.1 Overview

Figure 1 presents the proposed audio–text architecture. Given a video, the model processes its full transcript and speech waveform through two modality-specific pre-trained encoders. The resulting token-level linguistic representation and frame-level acoustic representation are projected into a common embedding space and passed through two bidirectional cross-attention blocks. The cross-attended sequences are then aggregated by masked mean pooling. Finally, the pooled representations and their pairwise interactions are combined to predict whether ambivalence or hesitancy (A/H) is present in the video.

### 4.2 Problem Formulation and Input Processing

For the $i$-th video, let $x_i^t$ denote its transcript, $x_i^a$ its audio waveform, and $y_i \in \{0, 1\}$ its video-level label. The label $y_i = 0$ represents No A/H, whereas $y_i = 1$ indicates the presence of A/H. The objective is to learn a model $g_\theta$ that

estimates

$$\mathbf{p}_i = p_\theta\left(y_i \mid x_i^t, x_i^a\right) = softmax\left(g_\theta(x_i^t, x_i^a)\right), \tag{1}$$

where $\mathbf{p}_i = [p_{i,0}, p_{i,1}] \in \mathbb{R}^2$ contains the posterior probabilities of the two classes. The hard prediction is

$$\hat{y}_i = \arg\max_{c \in \{0,1\}} p_{i,c}. \tag{2}$$

The transcript is tokenized with the cased multilingual BERT tokenizer and truncated to a maximum of 256 tokens. The waveform is converted to mono and resampled to 16 kHz. Recordings longer than 30 s are cropped: random cropping is applied during training, while centre cropping is used during validation and testing. Shorter sequences are retained in full. Text and audio sequences are padded within each mini-batch, and binary masks identify valid positions throughout the model.

### 4.3 Modality-Specific Encoding

The text branch uses the pre-trained bert-base-multilingual-cased encoder [?]. Given a tokenized transcript of length $N_t$, it produces contextual token representations

$$\mathbf{H}^t = mBERT(x_t) \in \mathbb{R}^{N_t \times d_t}, \qquad d_t = 768. \tag{3}$$

The corresponding text mask is denoted by $\mathbf{m}^t \in \{0, 1\}^{N_t}$.

The audio branch uses the pre-trained facebook/wav2vec2-base encoder [?]. For a waveform $x^a$, wav2vec 2.0 first applies its convolutional feature encoder and then contextualizes the downsampled acoustic sequence using its Transformer encoder:

$$\mathbf{H}^a = Wav2Vec2(x_a) \in \mathbb{R}^{N_a \times d_a}, \qquad d_a = 768, \tag{4}$$

where $N_a$ is the number of downsampled acoustic frames. The raw waveform mask is converted to the corresponding feature-level mask $\mathbf{m}^a \in \{0, 1\}^{N_a}$. In the reported configuration, both pre-trained encoders are frozen, and the subsequent projection, fusion, and classification layers are optimized.

Since the two encoders operate in different representation spaces, their outputs are independently projected into a shared $d = 256$ dimensional space. For modality $r \in \{t, a\}$, the projection is

$$P_r(\mathbf{H}^r) = Dropout(GELU(LN(\mathbf{H}^r\mathbf{W}_r + \mathbf{b}_r))), \tag{5}$$

where $\mathbf{W}_t \in \mathbb{R}^{768\times256}$ and $\mathbf{W}_a \in \mathbb{R}^{768\times256}$. The projected sequences are

$$\mathbf{T}_0 = P_t(\mathbf{H}_t) \in \mathbb{R}^{N_t\times256}, \qquad \mathbf{A}_0 = P_a(\mathbf{H}_a) \in \mathbb{R}^{N_a\times256}. \tag{6}$$

Layer normalization and GELU are used before dropout, whose probability is set to 0.2.

### 4.4 Bidirectional Cross-Attention Fusion

The projected sequences have different lengths and no explicit token-to-frame alignment. We therefore use bidirectional cross-attention to learn their relationships directly. The fusion module contains $L = 2$ cross-attention blocks, each with $H = 4$ heads. Given the input sequences $\mathbf{T}_{l-1}$ and $\mathbf{A}_{l-1}$ at block $l$, the text-to-audio direction uses text features as queries and acoustic features as keys and values. For head $h$, we compute

$$\mathbf{Q}_{l,h}^{t\leftarrow a} = \mathbf{T}_{l-1}\mathbf{W}_{l,h}^{Q,t},$$

$$\mathbf{K}_{l,h}^{t\leftarrow a} = \mathbf{A}_{l-1}\mathbf{W}_{l,h}^{K,a},$$

$$\mathbf{V}_{l,h}^{t\leftarrow a} = \mathbf{A}_{l-1}\mathbf{W}_{l,h}^{V,a}.$$

The attended context is

$$\mathbf{C}_{l,h}^{t\leftarrow a} = softmax\left(\frac{\mathbf{Q}_{l,h}^{t\leftarrow a}(\mathbf{K}_{l,h}^{t\leftarrow a})^\top}{\sqrt{d_h}} + \mathbf{M}^a\right)\mathbf{V}_{l,h}^{t\leftarrow a}, \tag{7}$$

where $d_h = d/H = 64$. The additive mask $\mathbf{M}^a$ is zero at valid acoustic positions and $-\infty$ at padded positions. Outputs from all heads are concatenated and linearly transformed:

$$\mathbf{C}_l^{t\leftarrow a} = Concat_{h=1}^{H}\left(\mathbf{C}_{l,h}^{t\leftarrow a}\right)\mathbf{W}_l^{O,t}. \tag{8}$$

The reverse audio-to-text direction uses acoustic features as queries and text features as keys and values:

$$\mathbf{C}_l^{a\leftarrow t} = MHA\left(\mathbf{A}_{l-1}, \mathbf{T}_{l-1}, \mathbf{T}_{l-1}; \mathbf{m}^t\right). \tag{9}$$

Equations eq:text$_{audio_m}$haandeq : audio$_t$ext$_m$haareevaluatedfromthesameblockinputs.Hence, thetextsequenceretrievesrelev

Each direction is followed by dropout, a residual connection, and layer normalization:

$$\hat{\mathbf{T}}_l = LN\left(\mathbf{T}_{l-1} + Dropout(\mathbf{C}_l^{t\leftarrow a})\right),$$
$$\hat{\mathbf{A}}_l = LN\left(\mathbf{A}_{l-1} + Dropout(\mathbf{C}_l^{a\leftarrow t})\right)$$

The two modalities then pass through separate position-wise feed-forward networks:

$$FFN_r(\mathbf{X}) = Linear_{1024\to256}(Dropout(GELU(Linear_{256\to1024}(\mathbf{X})))), \quad r \in \{t, a\}. \tag{10}$$

The output of block $l$ is

$$\mathbf{T}_l = LN\left(\hat{\mathbf{T}}_l + Dropout(FFN_t(\hat{\mathbf{T}}_l))\right),$$
$$\mathbf{A}_l = LN\left(\hat{\mathbf{A}}_l + Dropout(FFN_a(\hat{\mathbf{A}}_l))\right)$$

### 4.5 Masked Pooling and Interaction-Based Classification

After the final cross-attention block, the variable-length sequences are converted into fixed-dimensional representations by masked mean pooling. For the text modality,

$$\mathbf{t} = \frac{\sum_{j=1}^{N_t} m_j^t \mathbf{T}_{L,j}}{\max\left(1, \sum_{j=1}^{N_t} m_j^t\right)} \in \mathbb{R}^{256}, \tag{11}$$

and the acoustic representation is obtained analogously:

$$\mathbf{a} = \frac{\sum_{k=1}^{N_a} m_k^a \mathbf{A}_{L,k}}{\max\left(1, \sum_{k=1}^{N_a} m_k^a\right)} \in \mathbb{R}^{256}. \tag{12}$$

Rather than using simple concatenation alone, the classifier explicitly includes multiplicative agreement and absolute difference between modalities:

$$\mathbf{z} = [\mathbf{t}; \mathbf{a}; \mathbf{t} \odot \mathbf{a}; |\mathbf{t} - \mathbf{a}|] \in \mathbb{R}^{1024}, \tag{13}$$

where $[\cdot;\cdot]$ denotes concatenation and $\odot$ is element-wise multiplication. The product term represents coordinate-wise agreement, while the absolute difference preserves modality discrepancy, which may be informative when linguistic and vocal cues are inconsistent.

The interaction vector is processed by a two-layer fusion head:

$$\mathbf{u} = Dropout(GELU(LN(\mathbf{W}_1\mathbf{z} + \mathbf{b}_1))) \in \mathbb{R}^{256},$$
$$\mathbf{f} = Dropout(GELU(\mathbf{W}_2\mathbf{u} + \mathbf{b}_2)) \in \mathbb{R}^{128},$$
$$\mathbf{o} = \mathbf{W}_c\mathbf{f} + \mathbf{b}_c \in \mathbb{R}^{2},$$

where $\mathbf{o}$ contains the No-A/H and A/H logits. The posterior probabilities in Eq. eq:problem are obtained by applying softmax to $\mathbf{o}$.

### 4.6 Training Objective

The model is trained with class-weighted cross-entropy. For a mini-batch of $B$ samples, the implemented objective is

$$\mathcal{L} = -\frac{\sum_{i=1}^{B} w_{y_i} \log p_{i,y_i}}{\sum_{i=1}^{B} w_{y_i}}, \tag{14}$$

where the balanced weight of class $c$ is computed from the training set as

$$w_c = \frac{N}{CN_c}, \qquad C = 2, \tag{15}$$

with $N$ denoting the total number of training samples and $N_c$ the number of samples in class $c$. For the BAH training split, the resulting weights are approximately $w_0 = 0.990$ and $w_1 = 1.010$.

# 5 Experiments

## 5.1 Experimental Setup

We follow the subject-independent BAH protocol described in Section 3, using 778 videos for training, 124 for validation, and 525 for testing. Model selection is performed on validation Macro-F1, and the checkpoint with the highest validation score is used for test evaluation. All experiments use a fixed random seed of 42.

The model is trained for 10 epochs with a batch size of 4. We use AdamW with a learning rate of $3 \times 10^{-4}$, weight decay of 0.01, and gradient clipping with a maximum norm of 1.0. The learning rate follows a linear schedule with a warm-up ratio of 0.1. The multilingual BERT and wav2vec 2.0 encoders are frozen, while the projection layers, cross-attention blocks, fusion head, and classifier are optimized. The best validation Macro-F1 of 61.70% is obtained at epoch 4.

## 5.2 Evaluation Metrics

We report accuracy, precision, recall, and F1-score. A/H (label 1) is treated as the positive class when computing precision, recall, and F1-score. We additionally report Macro-F1, which averages the F1-scores of the No-A/H and A/H classes and therefore gives equal importance to both classes.

## 5.3 Results

Table 1reports the video-level results on the BAH test set. The proposed audio–text model achieves an accuracy of 72.19%, an A/H F1-score of 77.95%, and a Macro-F1 of 70.16%.

Table 1: Video-level performance on the BAH test set. Precision, recall, and F1-score are computed with A/H as the positive class. All values are percentages.

| Accuracy | Precision | Recall | F1-score | Macro-F1 |
|---|---|---|---|---|
| 72.19 | 75.00 | 81.13 | 77.95 | 70.16 |

Table 2presents the corresponding confusion matrix. The model correctly identifies 258 of the 318 A/H videos and 121 of the 207 No-A/H videos. Its recall is higher for A/H than for No A/H, indicating that the model is more sensitive to the presence of A/H but produces more false positives for the No-A/H class.

Table 2: Confusion matrix on the BAH test set.

| Ground truth | Predicted No A/H | Predicted A/H |
|---|---|---|
| No A/H | 121 | 86 |
| A/H | 60 | 258 |

# 6 Conclusion

We presented an audio–text framework for video-level ambivalence and hesitancy recognition. The model combines contextual representations from multilingual BERT and wav2vec 2.0 through stacked bidirectional cross-attention, followed by masked pooling and interaction-based classification. On the subject-independent BAH test set, the proposed model achieves 72.19% accuracy, 77.95% A/H F1-score, and 70.16% Macro-F1, demonstrating the potential ofjointly modelling linguistic content and acoustic behaviour for A/H recognition.

The current system uses only transcripts and speech, while the visual and fine-grained temporal information available in BAH remains unexplored. Future work will investigate unimodal and fusion ablations, adaptation of the pre-trained encoders, longer audio context, and the integration of visual cues to obtain a more complete representation of A/H behaviour.